\documentclass[sigconf]{acmart}
\usepackage{wasysym}
\usepackage{microtype}
\usepackage{multirow}
\usepackage[linesnumbered, ruled]{algorithm2e}
\usepackage{amsmath}
\usepackage{appendix}

\AtBeginDocument{%
 \providecommand\BibTeX{{%
 \normalfont B\kern-0.5em{\scshape i\kern-0.25em b}\kern-0.8em\TeX}}}
\begin{document}\sloppy

%\title{Counting-for-Tracking: Drone-based Crowd Tracking via Density-Aware Motion-Appearance Synergy}
\title[DenseTrack: Drone-based Crowd Tracking via Density-aware Motion-appearance Synergy]{DenseTrack: Drone-based Crowd Tracking via\\Density-aware Motion-appearance Synergy}

%% author
\author{Yi Lei}
\email{ly777@whut.edu.cn}
\authornote{Equal contribution}
\affiliation{%
 \institution{Wuhan University of Technology}
 \country{}
}

\author{Huilin Zhu}
\email{jsj_zhl@whut.edu.cn}
\authornotemark[1]
\affiliation{%
 \institution{Wuhan University of Technology and Singapore Management University}
 \country{}
}

\author{Jingling Yuan}
\email{yjl@whut.edu.cn}
\affiliation{%
 \institution{Wuhan University of Technology}
 \country{}
}

\author{Guangli Xiang}
\email{glxiang@whut.edu.cn}
\affiliation{%
 \institution{Wuhan University of Technology}
 \country{}
}

\author{Xian Zhong}
\email{zhongx@whut.edu.cn}
\authornote{Corresponding author}
\affiliation{%
 \institution{Wuhan University of Technology and Nanyang Technological University}
 \country{}
}

\author{Shengfeng He}
\email{shengfenghe@smu.edu.sg}
\affiliation{%
 \institution{Singapore Management University}
 \country{}
}

\begin{abstract}
Drone-based crowd tracking faces difficulties in accurately identifying and monitoring objects from an aerial perspective, largely due to their small size and close proximity to each other, which complicates both localization and tracking. To address these challenges, we present the Density-aware Tracking (DenseTrack) framework. DenseTrack capitalizes on crowd counting to precisely determine object locations, blending visual and motion cues to improve the tracking of small-scale objects. It specifically addresses the problem of cross-frame motion to enhance tracking accuracy and dependability. DenseTrack employs crowd density estimates as anchors for exact object localization within video frames. These estimates are merged with motion and position information from the tracking network, with motion offsets serving as key tracking cues. Moreover, DenseTrack enhances the ability to distinguish small-scale objects using insights from the visual-language model, integrating appearance with motion cues. The framework utilizes the Hungarian algorithm to ensure the accurate matching of individuals across frames. Demonstrated on \textsc{DroneCrowd} dataset, our approach exhibits superior performance, confirming its effectiveness in scenarios captured by drones. Our code will be available at: \url{https://github.com/Zebrabeast/DenseTrack}.

\end{abstract}

\begin{CCSXML}
<ccs2012>
 <concept>
 <concept_id>00000000.0000000.0000000</concept_id>
 <concept_desc>Computing methodologies, Tracking</concept_desc>
 <concept_significance>500</concept_significance>
 </concept>
 <concept>
 <concept_id>00000000.00000000.00000000</concept_id>
 <concept_desc>Human-centered computing, Collaborative and social computing</concept_desc>
 <concept_significance>300</concept_significance>
 </concept>
 <concept>
 <concept_id>00000000.00000000.00000000</concept_id>
 <concept_desc>Computing methodologies, Computer vision</concept_desc>
 <concept_significance>100</concept_significance>
 </concept>
 <concept>
 % <concept_id>00000000.00000000.00000000</concept_id>
 % <concept_desc>Do Not Use This Code, Generate the Correct Terms for Your Paper</concept_desc>
 % <concept_significance>100</concept_significance>
</ccs2012>
\end{CCSXML}

\ccsdesc[500]{Computing methodologies~ Tracking}
\ccsdesc[300]{Human-centered computing~ Collaborative and social computing}
\ccsdesc{Computing methodologies~ Computer vision}

\keywords{Multi-object Tracking, Crowd Localization, Vision-language Pre-training, Motion-appearance Fusion}

% \includegraphics[width = 0.99\textwidth]{sampleteaser}
% \caption{\small Seattle Mariners at Spring Training, 2010.}
% \Description{Enjoying the baseball game from the third-base
% seats. Ichiro Suzuki preparing to bat.}
% \label{fig:teaser}


\maketitle

\begin{figure}[t]
 	\centering
 	\includegraphics[width = \linewidth]{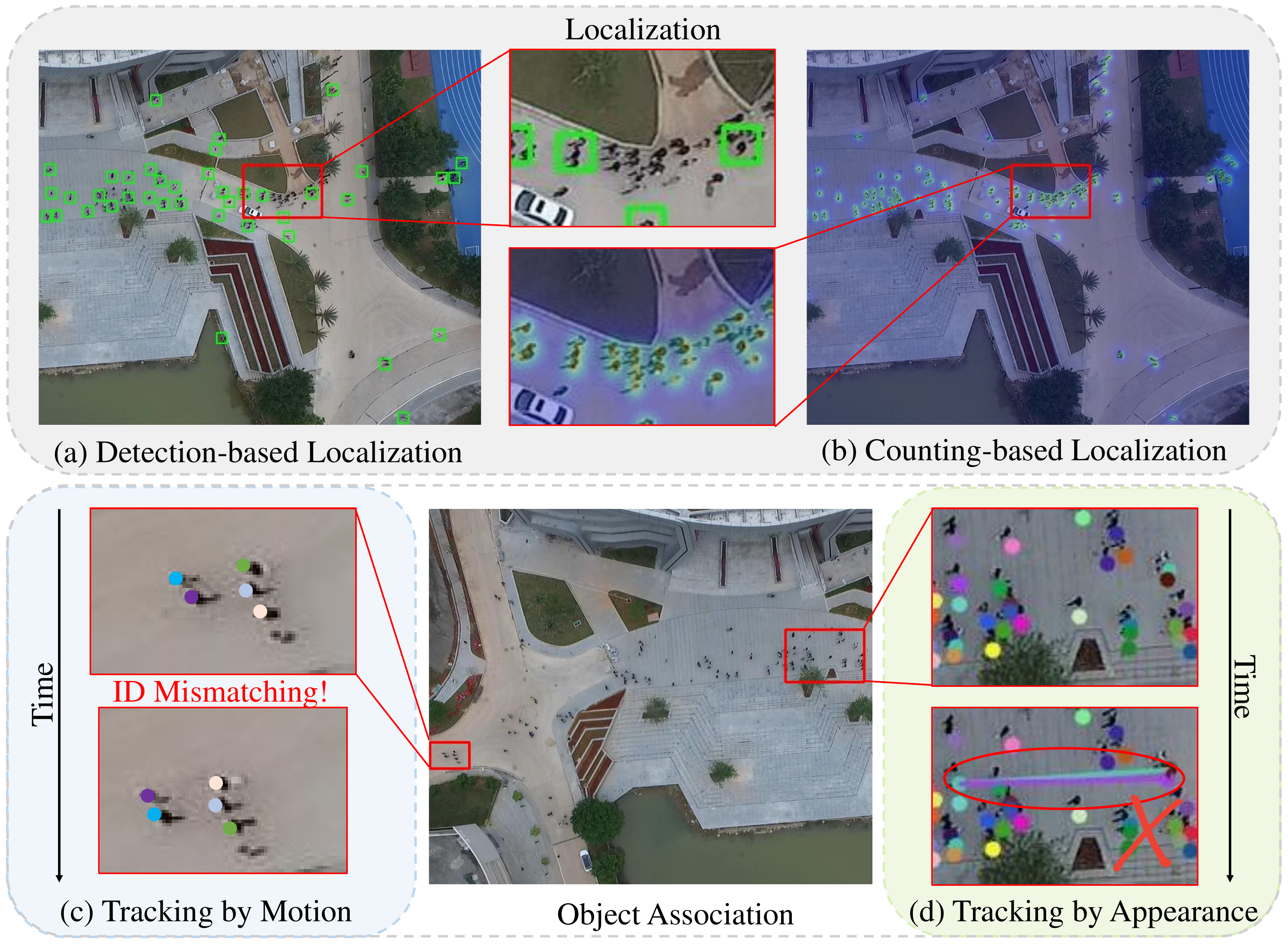}
 	\caption{\small Illustration of localization and tracking techniques. The upper section contrasts (a) detection-based localization, which relies on identifying objects directly, with (b) counting-based localization, which estimates object positions through density analysis. The lower section highlights inaccuracies in (c) Tracking by Motion, where predictions are based on movement patterns, and (d) Tracking by Appearance, which uses visual features; identically colored points indicate predictions for the same individual.}
 	\label{fig1}
\end{figure}

\section{Introduction}

Drone-based crowd tracking leverages unmanned aerial vehicle (UAV) cameras for automated surveillance, playing a critical role in crowd management and monitoring. This technology is designed to identify and consistently track individuals across successive video frames, amidst ongoing movement of both the subjects and the background. To achieve this, the process of multi-object tracking (MOT) is utilized~\cite{bergmann2019tracking, fu2023denoising, luo2021multiple, cvpr/CaoPWKK23, cvpr/ShuaiBLMT21}, which involves two critical steps: localization and tracking. Localization discerns the exact positions of objects within each frame, while tracking maintains consistent identification of these objects over time. These tasks are complicated by factors such as object size, density, and environmental complexity. Fig.~\ref{fig1} depicts the various approaches to localization and tracking, showing the challenges each method faces.

Regarding the localization task, Figs.~\ref{fig1}(a) and (b) depict the performance of detection-based and counting-based methods. Detection methods struggle with small objects and complex backgrounds, often resulting in significant errors. Conversely, counting-based methods provide almost accurate localization of all target individuals in densely populated scenes. However, unlike detection methods, counting approaches sacrifice a considerable amount of individual appearance information, which complicates the use of similarity-based tracking techniques in MOT with bounding boxes. This loss presents a substantial challenge in balancing precise localization with the preservation of appearance information for individuals.

For the tracking task, as shown in Figs.~\ref{fig1}(c) and (d), the methods of tracking by motion and tracking by appearance are explored. Tracking by motion effectively considers the inter-frame movement of objects but can lead to misidentification in scenarios with dense, small targets. On the other hand, tracking by appearance, while focusing solely on the visual attributes, often mistakenly associates distant and different objects due to ignoring inter-frame motion. Thus, the second challenge involves effectively using inter-frame physical distances to minimize these errors.

In addressing the first challenge, various counting-based tracking methods have been developed to balance precise localization with the preservation of appearance information for individuals from drone perspectives. For instance, STNNet~\cite{cvpr/WenDZHWBL21} leveraged density maps for crowd localization and motion offsets for tracking. Although this method significantly enhances localization accuracy, it struggles with object displacement issues, particularly due to the small size and close proximity of objects in aerial views. Additionally, the multi-frame attention-based method~\cite{wacv/AsanomiNB23} aims to improve tracking by integrating features across multiple frames. However, its dependency on consecutive frames reduces its effectiveness in scenarios characterized by large inter-frame intervals.

Addressing the second challenge involves MOT methods~\cite{icip/MaggiolinoACK23, tmm/DuZSZSGM23, iccv/CuiZZYWW23, cvpr/SeidenschwarzBS23} that combine appearance and motion cues to capitalize on their strengths and mitigate errors. However, tracking in drone-based environments presents unique difficulties, particularly with the detection of small objects. Extracting individual appearance features from density maps in drone perspectives is notably challenging and less effective compared to detection-based methods, which inherently capture richer detail, whereas density maps offer limited information.

In this paper, we introduce the Density-aware Tracking (\textit{DenseTrack}) framework, which advances the counting-based localization framework by incorporating both motion and appearance cues. DenseTrack tackles two critical tasks: extracting detailed appearance information from density maps for precise individual identification and correcting motion discrepancies using this appearance data. Initially, DenseTrack utilizes visual-language models (VLMs) to derive intricate appearance features from density maps, ensuring accurate characterizations of individuals. The appearance data thus extracted is then seamlessly integrated with motion and position data to address motion inaccuracies, enhancing the fidelity of motion cues. This strategic integration effectively surmounts the challenges of object localization in drone-based scenarios, while adeptly merging both motion and appearance information into the tracking process.

In summary, our contributions to the field are threefold:

\begin{itemize} 
% \vspace{-\baselineskip}
\item We introduce the Density-aware Tracking (DenseTrack) framework, a novel approach that synergistically combines motion and appearance cues within a crowd counting localization paradigm. This strategy effectively exploits the strengths of both cues while mitigating their limitations.

\item We enhance the process of individual identification within density maps by integrating a visual-language model. This integration significantly improves the descriptive capabilities of density maps, enabling more nuanced and accurate representations of individuals in crowded scenes.

\item We demonstrate the superior performance of our approach using \textsc{DroneCrowd}, where it outperforms existing methods in the field of crowd tracking.
\end{itemize}

\section{Related Work}
\textbf{Crowd Counting}
is essential for effective crowd management and has received significant attention in recent years. It can be broadly classified into three categories: detection-based methods~\cite{arteta2014interactive}, regression-based methods, and density map-based methods~\cite{yan2021crowd, wang2022stnet, mm/ZhuYZYWH23}. While detection-based methods struggle in densely populated scenes, regression-based approaches often fail to localize individuals accurately in sparser crowds, making density map-based methods the preferred technique. This approach has proven superior to traditional methods, demonstrating exceptional efficacy. Previous solutions, such as multi-branch networks~\cite{cvpr/ZhangZCGM16}, aimed to address the varying scales of crowd distribution but typically produced suboptimal density maps. The introduction of null convolution has revolutionized this area by preserving pixel information and reducing parameter count, thereby enhancing performance. The evolution of deep learning has further expanded and improved the architecture of backbone networks. The strategic development of convolutional neural networks (CNNs) and the incorporation of Transformer networks into single-domain approaches have become increasingly effective~\cite{iccvw/SajidCSKW21, corr/abs-2109-14483, chinaf/LiangCXZB22, raaj2019efficient}. Recent innovations have even enabled precise crowd localization~\cite{icassp/HanGYW20, tmm/LiangXZZ23,xu2020segment, tmm/ZhuYZLW24}. Despite these advancements, solely focusing on crowd numbers is insufficient for comprehensive crowd management. Assessing crowd movement is also important for identifying potential risks within a crowd.

\textbf{Multi-object Tracking}
%(MOT)~\cite{zhao2017DenseTracker} 
poses a considerable challenge in computer vision, involving the detection and continuous tracking of multiple objects across video sequences~\cite{tcsv/SunCLRM21,zhou2022global, wang2022deformable, ijon/SunCMLRP22}. Traditionally, methods such as active contours~\cite{tip/HuZLLZM13}, particle filters~\cite{tip/ZhangLXLY18}, and various association techniques~\cite{voigtlaender2019mots, eccv/ZhangSJYWYLLW22, icip/ZhongTRHXY20} have been employed. However, there has been a significant shift towards a tracking-by-detection paradigm in recent years. This approach uses bounding-box detectors to identify objects and leverages appearance features for association, although it often struggles with accurately detecting smaller objects due to their lack of distinctive features.

The Simple Online Real-Time Tracker~\cite{bewley2016simple} provided an efficient solution for MOT, featuring rapid update frequencies and minimal processing requirements. Building on SORT, DeepSort~\cite{icip/WojkeBP17} incorporated deep learning-based association metrics to significantly enhance tracking accuracy by using more sophisticated data association techniques. Additionally, \citet{eccv/ZhangSJYWYLLW22} developed ByteTracker, an advanced tracking algorithm that utilizes deep neural networks. ByteTracker is noted for its exceptional accuracy and robust performance in challenging environments, making it a powerful tool for complex MOT tasks.

\textbf{Crowd Tracking}
has witnessed significant advancements, with innovative developments reshaping the field. \citet{cvpr/KratzN10} utilized a space-time model to track individuals within crowds effectively. AdaPT~\cite{icra/BeraGSLM14} introduced a real-time algorithm that deduces individual trajectories in dense environments, enhancing the understanding of crowd dynamics. Recent methods such as tracking-by-counting~\cite{tip/RenWTTC21} integrated detection, counting, and tracking to leverage complementary data, proving to be effective for real-time people counting applications~\cite{tits/SunASZLM19}. Furthermore, \citet{cvpr/SundararamanBMP21} developed the Congested Heads Dataset, which combines a head detector with a Particle Filter and a re-identification module to efficiently track multiple individuals in crowded settings.

\begin{figure*}[t]
	\centering
	\includegraphics[width = 0.99\textwidth]{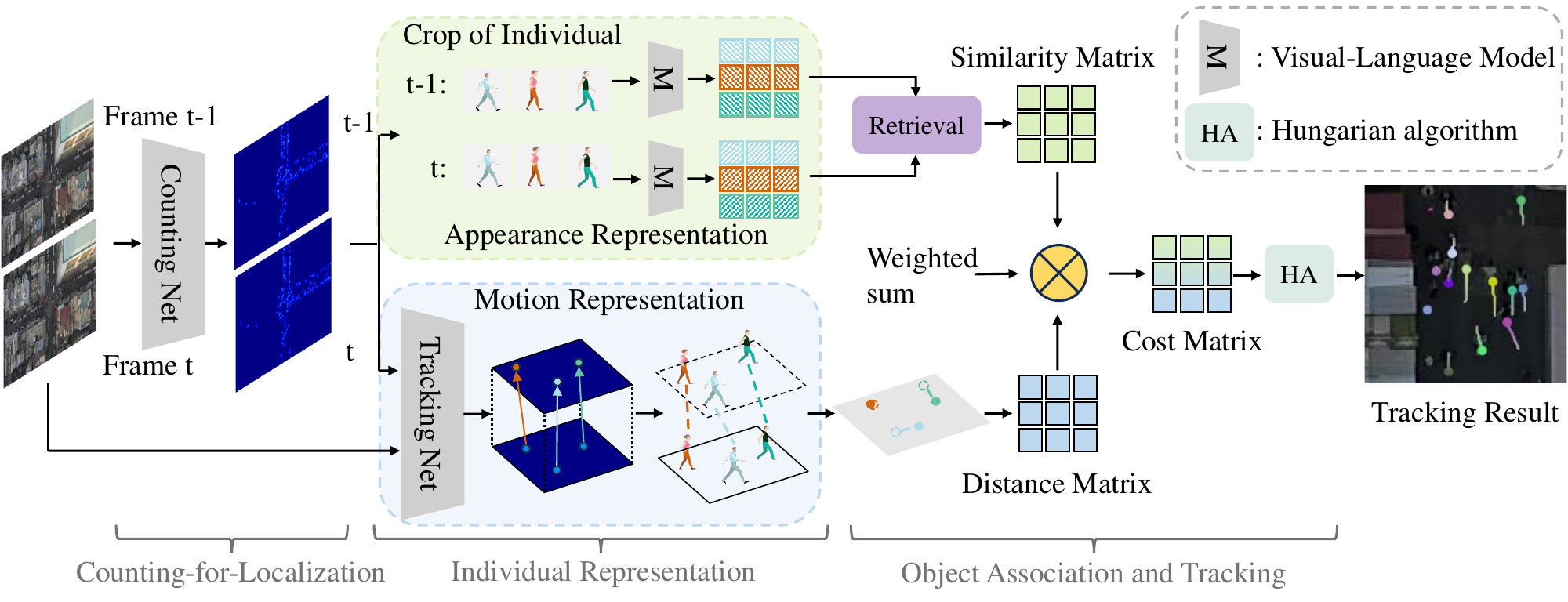}
	\caption{\small DenseTrack is structured around three essential components: Localization, Individual Representation, and Association. Localization accurately determines the spatial positions of individuals in crowds through density maps. For Individual Representation, motion and appearance features are extracted by aligning density maps with motion and position maps (MPM) to provide motion cues, while the BLIP2 method is used to gather appearance cues. The Association component employs diffusion-based retrieval alongside a distance matrix derived from motion cues to facilitate precise inter-frame individual matching.}
	\label{fig2}
\end{figure*}

\section{DenseTrack}

\subsection{Problem Formulation}
% 图2概述了所建议的方法。
Focusing on small, densely packed objects, this paper introduces a counting-based method for drone-based crowd tracking, integrating appearance and motion cues to compensate for their respective limitations.
The framework, depicted in Fig.~\ref{fig2}, comprises three stages: Localization, Individual Representation, and Object Association and Tracking.
The input consists of all frames $I = \{I_1, I_2, \cdots, I_N\}$ from the video stream $V$, where $N$ denotes the total number of frames. The output includes trajectories $T = \{T_1, T_2, \cdots, T_M\}$ for each individual within the video stream $V$, with $M$ denoting the total number of individuals detected.

%, which will be detailed in Sec.~\ref{sec.3.2}, Sec.~\ref{sec.3.3}, and Sec.~\ref{sec.3.4}, respectively.

% 在Localization阶段, 将视频流V的所有帧I = {I_1, I_2, ..., I_N}逐帧输入到人群计数网络（CN）中, 以得到每帧图像对应的密度图D = {D_1, D_2, ..., D_N}, 如式（1）所示。

The Localization stage involves sequentially inputting all frames $I$ from the video stream $V$ into the crowd counting network ($\mathrm{CN}$) to derive the coordinate list $\mathrm{CL} = \{\mathrm{CL}_1, \mathrm{CL}_2, \cdots, \mathrm{CL}_N\}$ for each frame image, given by:
\begin{equation}
	\mathrm{CL}_i = \mathrm{CN} \left(I_i \right), \quad \left(0 \le i < N \right).
 \label{eq1}
\end{equation}

In the Individual Representation (IR) stage, all frames $I$ from the video stream $V$, along with the coordinate list $\mathrm{CL}i$ of individuals in each frame $I_i$, are inputted. 
Then, leveraging the localization from density maps, we obtain estimated positions $\tilde{\mathrm{CL}}_{i-1}$ from the last frame $I_{i-1}$ and appearance representations $F_i$ of individuals in each frame $I_i$. The formula is defined as:
\begin{equation}
	\tilde{\mathrm{CL}}_{i-1}, \hat{F_{i}} = \mathrm{IR} \left(I_i, \mathrm{CL}_i \right). 
 \label{eq2}
\end{equation}

In the final stage, Object Association and Tracking (OAT), individuals' appearance representations $F = \{F_1, F_2, \cdots, F_N\}$ function as appearance cues, while the estimated positions $\tilde{\mathrm{CL}} = \{\tilde{\mathrm{CL}}_1, \tilde{\mathrm{CL}}_2, \cdots, \tilde{\mathrm{CL}}_{N-1}\}$ and coordinate list $\mathrm{CL}$ serve as motion cues. This stage entails matching individuals across different frames, culminating in the derivation of individual trajectories $T$ as:
\begin{equation}
	T = \mathrm{OAT} \left(\tilde{\mathrm{CL}}, \mathrm{CL}, \hat{F} \right). 
 \label{eq3}
\end{equation}

\subsection{Localization}
\label{sec.3.2}
% 由于检测器对于无人机高空俯视视角下的小目标并不擅长, 所以我们引入了人群计数网络定位, 替换了常用的detection网络。具体地, 我们逐帧递入视频流V的所有帧, 以获得其对应的密度图D。
% Since detectors are not proficient at detecting small objects from the high-altitude overhead perspective of drones, we introduced crowd counting network localization as a replacement for commonly used detection networks. 
Localization forms the basis of tracking. Given detectors' limitations in identifying small objects from the high-altitude overhead perspective of drones, it's essential to establish a solid tracking foundation. Thus, we introduce crowd counting network localization as a replacement for traditional detection networks.
Specifically, we input all frames $I$ of the video stream $V$ frame by frame to obtain their corresponding density maps $D = \{D_1, D_2, \cdots, D_N\}$.

% 但是, 被广泛使用的密度图在密集区域存在高斯斑点重合的问题, 不利于进行准确的人群定位。受Liang等人工作的启发, 我们使用Focal Inverse Distance Transform MAP作为密度图, 使用High-Resolution Network（HR）对密度图进行预测, 如式4所示。
However, the prevalent issue with widely used density maps, lacking precise individual localization, impedes accurate crowd localization.
Inspired by~\cite{tmm/LiangXZZ23}, we employ the focal inverse distance transform map as the density map and use the high-resolution network ($\mathrm{HR}$) for density map prediction:

\begin{equation}
	D_i = \mathrm{HR} \left(I_i \right), \quad \left(0 \le i < N \right). 
 \label{eq4}
\end{equation}

% 在得到了视频逐帧的密度图后, 密度图每个像素点上的值对应着该点存在人的可能性, 因此, 如果密度图上的某个点式局部最大值, 那么我们则认为该点的坐标是帧中某个个体的坐标, 由此, 我们可以得到每帧中所有个体的坐标CL_i。
After obtaining the density map $D$ for each frame of the video, each pixel in the density map signifies the likelihood of an individual's presence. 
Consequently, in this density-aware stage, if a point on the density map is a local maximum ($\mathrm{LM}$), the coordinates of that point are considered as the coordinates of an individual in the frame. Thus, the coordinate list of all individuals in each frame is derived, denoted as $\mathrm{CL}_i$:
\begin{equation}
	\mathrm{CL}_i = \mathrm{LM} \left(D_i \right), \quad \left(0 \le i < N \right).
 \label{eq5}
\end{equation}

\subsection{Individual Representation}
\label{sec.3.3}
% 在得到了每帧中个体的准确位置后, 我们需要提取足够且有效的个人表示来用于个体的帧间关联。运动特征与外观特征在通常情况下都是可以单独作为帧间关联的线索的。但是在密集的情况下, 仅仅使用运动偏移量作为个人表示进行帧间关联容易在聚集的人群中失效。而在无人机的高空视角下, 人的外观特征并不明显, 也不足以单独作为帧间关联的依据。因此, 我们同时提取外观特征和运动特征, 作为帧间关联的线索。
After obtaining accurate positions of individuals in each frame, extracting effective representations for inter-frame association is crucial. To integrate both appearance and motion information, the simultaneous extraction of both appearance features $F$ and motion offsets $\vec{o}$ is adopted as association cues.

\subsubsection{Appearance Representation}
% 为了克服高空俯视视角下难以获取具有区别性的外观特征的难题, 需要使用一个强大的外观提取器, 来获取帧中所有个体的外观表示$F$。
Considering the inherent limitations of density maps in providing detailed individual information and the critical role of rich appearance features in tracking accuracy, we proceed to acquire appearance representations $F$ for all individuals in the frames.

For unsupervised extraction of individual representations, we employ the vision-language pre-training model, BLIP2~\cite{Li2023BLIP2}.
By utilizing $\mathrm{Cut}$, the original images are cropped based on individual localization, extracting local patches representing individuals in each frame. Denoted as $\mathrm{cl}_{i, j} \in \mathrm{CL}i (0 \le j < M_i)$, where $M_i$ represents the number of individuals appearing in the $i$-th frame, these individuals are then used to obtain sub-images $S{i, j}$ corresponding to each individual:

\begin{equation}
	S_{i, j} = \mathrm{Cut} \left(\mathrm{cl}_{i, j}, I_i \right).
\label{eq6}
\end{equation}
% 在得到了每帧中每个个体的sub-img后, 我们使用BLIP的特征提取模块（BE）, 获得每帧中每个个体的外观表示, 公式定义如式六所示。
After obtaining individual local patches $S_{i, j}$ for each individual in every frame, BLIP2's feature extraction ($\mathrm{BE}$) module is employed to acquire appearance representations $F_{i, j}$ for each individual:
\begin{equation}
	F_{i, j} = \mathrm{BE} \left(S_{i, j} \right), 
 \label{eq7}
\end{equation}
where the representation $\hat{F_{i, j}}$ obtained here is a matrix with dimensions $(W, H)$, which is not convenient for merging all individual identifiers in subsequent frames. Therefore, we flatten the matrix $F_{i, j}$ to obtain $\hat{F_{i, j}}$, with dimensions $(1, W \times H)$.

\subsubsection{Motion Representation}
% 在多目标跟踪任务中, 运动偏移量一直都作为运动表示用于帧间关联。为了准确获取密集小目标场景中的个体的运动偏移量, 我们必须基于密度图对于帧中个体的精确定位$CL_{}$来获取对应位置的运动信息, 但是密度图本身只能进行计数和定位, 并不能提供帧中个体的运动偏移量。因此, 我们需要使用一个方法通过个体的定位, 获取个体的运动信息。
In the Motion Representation stage, accurately determining the motion offsets of individuals in dense small object scenes is crucial. We utilize density maps to locate individuals within the frame $\mathrm{CL}_{i, j}$ and gather the corresponding motion information at these positions.
However, density maps lack motion offsets $\vec{o}_{i, j}$ and are limited to counting and localization. Thus, motion information extraction is necessary as density maps lack individual motion offsets.
% 受~\cite{cvpr/HayashidaNB20}的成功的启发, 我们也使用motion and position map(MPM)对于个体的运动状态进行预测。具体来说, 在MPM中, 每个个体$j$所在的像素点上的值$C_{i-1, i}^j$, 是基于该个体的运动偏移量进行计算的, 如式8所示。

Inspired by~\cite{cvpr/HayashidaNB20}, we utilize the motion and position map (MPM) to predict the motion states of individuals. Specifically, given frames $i$ and $i+1$, we generate MPM $C_{i+1}$. In MPM $C_{i+1}$, the value $C_{{i+1}, j}$ at each pixel point where individual $j$ is located is calculated based on the motion offset of that individual as follows:
\begin{equation}
	C_{{i+1}, j} = G \left(p_t \right) \frac{\mathrm{cl}_\mathrm{i, j} - \mathrm{cl}_{{i+1}, j}}{\left\| \mathrm{cl}_\mathrm{i, j} - \mathrm{cl}_{{i+1}, j} \right\|_2}, 
 \label{eq8}
\end{equation}
where $G(p_t)$ is specifically derived through Gaussian filtering and signifies the likelihood that the point corresponds to an individual.

% 因此, 输入第$i$帧和第$i+1$帧可以得到预测的MPM图$\tilde{C}_{}$。公式如式14定义。
% Therefore, by inputting frames $I_i$ and $I_{i+1}$ into Tracking Net (TN), we can obtain the predicted MPM $\tilde{C}_{i, {i+1}}$. The formula is defined as:
Then, frames $I_i$ and $I_{i+1}$ are inputted into the Tracking Net (TN) to generate the motion and position map (MPM) $\tilde{C}_{i, i+1}$ using the following formula:
\begin{equation}
\tilde{C}_{i+1} = \mathrm{TN} \left(I_i, I_{i+1} \right),
\label{eq14}
\end{equation}
%这里MPM图$\tilde{C}_{}$实际上可以看成是一个W×H的矩阵, 如式15所示, W_I代表的是帧I的宽, H_I代表的是帧I的高, 而矩阵中的值是一个用于描述运动偏移量的向量, 其意义如式16所示。
where MPM $\tilde{C}_{i+1}$ is a matrix with the shape of $(w, h)$, where $w$ represents the width of frame $I$ and $h$ represents the height of frame $I$. The values in the matrix represent a vector $\vec{o}$ describing the motion offset.
\begin{equation}
	\tilde{C}_{i+1} = \begin{bmatrix}
	\vec{o}_{1, 1} & \cdots & \vec{o}_{1, w}\\
	\vdots &\ddots &\vdots\\
	\vec{o}_{h, 1} & \cdots & \vec{o}_{h, w}
	\end{bmatrix}.
	\label{eq15}
\end{equation}
%假设我们需要获取的是第i+1帧中第j个人在第i帧和第i+1帧的运动偏移量, 那么我们只需要在$\tilde{C}_{}$中的对应位置获取相应的运动偏移量即可。由运动偏移量和坐标, 我们可以得到第i+1帧中的第j个人在第i+1帧中的估计位置$\tilde{P}{i, j}$, 如式9所示。
To obtain the motion offset of the $j$-th individual in the $(i+1)$-th frame relative to the $i$-th frame, we retrieve the corresponding motion offset $\vec{o}_{x, y}$ at the respective position in $\tilde{C}_{i+1}$. Using the motion offset and coordinates, we can calculate the estimated position $\tilde{\mathrm{cl}}_{i, j}$ of the $j$-th individual in the $(i+1)$-th frame as follows:
\begin{equation}
	\tilde{\mathrm{cl}}_{i, j} = \mathrm{cl}_{i+1, j} + \vec{o}_{x, y},
 \label{eq9}
\end{equation}
where $x$ and $y$ respectively represent the horizontal and vertical coordinates stored in $\mathrm{cl}_{i, j}$, serving as indices to retrieve the motion offset stored in $\tilde{C}_{i, {i+1}}$.

\subsection{Object Association and Tracking}
\label{sec.3.4}

% 多目标跟踪的本质实际上是对多个目标进行检测并赋予每个目标独一无二的ID进行轨迹跟踪。在之前的步骤中, 我们已经得到了每个人在每一帧中的位置, 完成了对多个目标进行检测的步骤, 因此多目标跟踪的任务转换为前后两帧之间的目标关联问题。
This paper focuses on MOT, which involves detecting multiple targets and assigning unique identities for trajectory tracking. After acquiring the positions of each individual in every frame, the task shifts to associating targets between consecutive frames. To enhance tracking accuracy, we integrate motion offsets and appearance features for inter-frame association.
% This paper focuses on the detection of multiple targets and assigning each a unique identity (ID) for trajectory tracking, which constitutes a fundamental aspect of Multiple Object Tracking (MOT). With the positions of each individual acquired in every frame, the preceding steps complete the detection of multiple targets. Subsequently, the task shifts to associating targets between consecutive frames.

% % 为了能够提高跟踪的准确性, 我们将运动偏移量和每个个体的外观特征同时作为帧间关联的依据。
% To enhance tracking accuracy, we concurrently consider motion offsets and the appearance features of each individual as the basis for inter-frame association.

% 具体来说, 如果第i帧中的第j个人与第i+1帧中的第k个人是同一个人, 那么我们认为, 它们的的外观是一定相似的。这个外观特征的对比过程, 与图片检索类似。因此, 受~\cite{yang2019efficient}工作的启发, 我们在外观特征关联阶段也使用Diffusion的方法, 将连续两帧中所有个体的外观特征$F_i$表示进行对比, 以获得关于外观的相似度矩阵$M^S_{\}$。
Specifically, in the appearance feature association stage, we consider the appearances of the $k$-th individual in frame $i$ and the $j$-th individual in frame $i+1$ to be inherently similar if they correspond. Drawing inspiration from the success of~\cite{yang2019efficient}, we utilize the diffusion method ($\mathrm{DM}$) to compare appearance representations $F_i$ across frames, akin to image retrieval. 
This process yields the similarity matrix $A^S_{i, i+1} \in \mathbb{R}^{p \times q}$, where $p$ represents the number of individuals detected in the previous frame $I_i$, and $q$ denotes the number of individuals detected in the subsequent frame $I_{i+1}$.
\begin{equation}
	A^S_{i, i+1} = \mathrm{DM} \left(F_i, F_{i+1} \right),
 \label{eq10}
\end{equation}
% 同时, 第i+1帧中出现的每个人被估计在第i帧中的位置应该与它们在第i帧中的实际位置$CL$是接近的, 因此, 第i+1帧中每个个体被估计在第i帧中的位置与第i帧中的每个个体的实际位置也可以构成一个关于预测位置与实际位置的矩阵$M^D_{i, i+1} \in \mathbb{R}^\mathrm{M_i \times M_{i+1}}$, 如式11所示。
where the values in $A^S_{i, i+1}$ are represented as follows:
\begin{equation}
	A^S_{i, i+1} = \{a^S_{k, j}\} = \begin{bmatrix}
	a^S_{1, 1} & \cdots & a^S_{1, q}\\
	\vdots & \ddots & \vdots\\
	a^S_{p, 1} & \cdots & a^S_{p, q}
	\end{bmatrix}, 
	\label{eq11}
\end{equation}
where $a^S_{k, j}(k\in(1,p),j \in (1,q))$ represents the appearance similarity score between the $k$-th individual in frame $i$ and the $k$-th individual in frame $i+1$, ranging between 0 and 1.

\begin{algorithm}[t]
\caption{\small Inter-Frame Association for Tracking}\label{algfortracking}
\footnotesize
\SetAlgoLined
\SetKwInOut{Input}{Input}
\SetKwInOut{Output}{Output}

\Input{Localization of individuals in each frame, $\mathrm{CL} = \{\mathrm{CL}_1, \mathrm{CL}_2, \cdots, \mathrm{CL}_N\}$; Cost matrix between two frames, $A^C_{i, i+1}$, for $1 \leq i < N$.}
\Output{Trajectories of individuals across frames, $T = \{T_1, T_2, \cdots, T_M\}$.}
\BlankLine
Initialize tracking for the first frame in the video:\\
\For{$j = 1$ \KwTo $\mathrm{Len}(\mathrm{CL}_1)$}{
 Assign initial positions: $T_{j, 1} = \mathrm{CL}_{1, j}$\\
}
\For{each subsequent frame $i = 1$ \KwTo $N$}{
 Compute matching pairs using the cost matrix:\\
 $\mathrm{ML}_{i-1, i} = \mathrm{HA}(A^C_{i-1, i})$\\
 \For{each match $k = 0$ \KwTo $\mathrm{Len}(\mathrm{ML}_{i-1, i})$}{
 Assuming the $id$-th trajectory is matched with the $u$-th individual in the $i$-th frame:\\
 Update trajectories: $T_{\mathrm{id}, i} = \mathrm{CL}_{i, \mathrm{u}}$\\
}
 If an individual $r$ in frame $i$ is unmatched, assign a new ID for a new trajectory:\\
 $\mathrm{id}_c = \mathrm{Len}(\mathrm{ML}_{i-1, i}) + 1$,\\
 $T_{\mathrm{id}_c, i} = \mathrm{CL}_{i, r}$\\
}

\end{algorithm}

Simultaneously, to ensure that the estimated positions $\tilde{P}_i$ of each individual appearing in frame $i+1$ closely align with their actual positions $\mathrm{CL}_i$ in frame $i$, we construct a matrix $A^D_{i, i+1} \in \mathbb{R}^{p \times q}$. This matrix represents the estimated positions of each individual in frame $i+1$ relative to the actual positions of individuals in frame $i$:
\begin{equation}
	A^D_{i, i+1} = \{a^D_{k, j}\}= \begin{bmatrix}
	a^D_{1, 1} & \cdots & a^D_{1, q}\\
	\vdots & \ddots &\vdots\\
	a^D_{p, 1} & \cdots & a^D_{p, q}
	\end{bmatrix}, 
	\label{eq12}
\end{equation}
% 这里的$m_{k, j}$表示的是第i帧中的第k个人的实际位置$CL_{i, k}$与第i+1帧中的第j个被预测在第i帧中的位置$\tilde{P}_{i, j}$之间的欧式距离：
where $a_{k, j}^D(k\in(1,p),j \in (1,q))$ represents the Euclidean distance between the actual position $\mathrm{CL}_{i, k}$ of the $k$-th individual in frame $i$ and the predicted position $\tilde{\mathrm{CL}}_{i, j}$ of the $j$-th individual in frame $i+1$, as estimated in frame $i$:
\begin{equation}
 a_{k, j}^D = \sqrt{\left(x - \tilde{x} \right)^2 + \left(y - \tilde{y} \right)^2}, 
 \label{eq13}
\end{equation}
%这里的${CL}^x_{i, k}$和${CL}^y_{i, k}$实际上是${CL}_{i, k}$在x轴上的坐标和在y轴上的坐标。$\tilde{P}^x_{i, j}$与$\tilde{P}^y_{i, j}$也是同理。
where $x$ and $y$ denote the coordinates of $\mathrm{CL}_{i, k}$ along the x- and y-axis, respectively, and similarly, $\tilde{x}$ and $\tilde{y}$ are defined.

% 从上述的步骤中, 我们可以知道, 无论是相似度矩阵$M^S_{i, i+1}$还是距离矩阵$M^D_{i, i+1}$, 它们中的值都可以用来作为衡量两帧中的个体是统一人的可能性的依据。但是, 如果仅靠相似度矩阵对结果进行匹配, 容易忽略距离问题, 而造成两帧中两个空间距离较远的人被赋予了同一ID。同时, 如果仅依靠距离矩阵对结果进行匹配, 聚集的人群将会因为有高度相似的距离线索而导致ID切换, 也不利于跟踪结果。
%因此, 我们采取运动和外观结合的方式进行帧间关联, 使这两个度量标准通过服务于分配问题的不同方面进行补充。具体来说, 我们首先将距离矩阵$M^D_{i, i+1}$中的值进行放缩, 使矩阵中的值在0到1之间, 我们将这样一个经过变换的距离矩阵称之为$M^D_{i, i+1}\prime$.

From the preceding steps, both the similarity matrix $A^S_{i, i+1}$ and the distance matrix $A^D_{i, i+1}$ offer means to gauge the likelihood that individuals in two frames are the same. 
However, if solely relying on the similarity matrix, distance issues are overlooked, potentially resulting in the assignment of the same ID to spatially distant individuals. Conversely, if matching relies solely on the distance matrix, ID switches can happen within clusters of individuals due to highly similar distance cues, harming tracking outcomes.

Therefore, a synergistic approach of motion and appearance for inter-frame association is adopted, aiming to complement these two metrics by addressing different aspects of the assignment problem. Initially, the values in the distance matrix $A^D_{i, i+1}$ are rescaled to range between 0 and 1, resulting in the transformed distance matrix denoted as $\hat{A}^D_{i, i+1}$.
% 为了构建关联问题, 我们使用加权和组合两个指标, 如式16所示。
To formulate the association problem, a weighted sum is employed to integrate both metrics, as follows:
\begin{equation}
 A^C_{i, i+1} = \left(-\lambda \right) \hat{A}^D_{i, i+1} + \left(1- \lambda \right) A^S_{i, i+1}.
 \label{eq16}
\end{equation}

%这个地方之所以要在距离矩阵$ M^D_{i, i+1}$前乘上$- \lambda $, 是因为在匹配任务中, 距离矩阵$ M^D_{i, i+1}$中的值是越小越成为一个人的可能性更高, 而相似度矩阵$ M^S_{i, i+1}$中的值则是越大越成为一个人的可能性更高。
Before combining the matrices, the distance matrix $A^D_{i, i+1}$ is multiplied by $-\lambda$ to adjust its influence. In the matching task, smaller values in $A^D_{i, i+1}$ suggest a higher likelihood of representing the same individual, while larger values in $A^S_{i, i+1}$ indicate a higher likelihood of representing different individuals.

%在得到了代价矩阵$M^C_{i, i+1}$之后, 我们利用匈牙利算法得到可以得到两帧之间在两个度量标准下最为匹配的结果, 将这些两帧之间的匹配结果链接起来, 就可以得到视频中每个人在每一帧中的轨迹 $T = \{T_1, T_2, \cdots, T_M\}$ , 具体过程如算法1所示。
After obtaining the cost matrix $A^C_{i, i+1}$, we employ the Hungarian algorithm (HA) to determine the optimal matches between frames using both metrics. This facilitates the establishment of associations across frames, enabling the deduction of each individual's trajectory in the video for every frame, denoted as $T$. The detailed procedure is outlined in Algorithm~\ref{algfortracking}.
Through the aforementioned operations, the trajectory $T$ is obtained, composed of the positions where each ID appears in every frame, completing the tracking process.

\section{Experimental Results}
% In this section, we will demonstrate the performance of our model on \textsc{DroneCrowd}, which is the only dataset specifically designed for tracking crowds captured by drones. Additionally, we conduct ablation experiments to validate the effectiveness of our modules.

\subsection{Dataset and Metrics}

Our experiments utilize \textsc{DroneCrowd} dataset~\cite{cvpr/WenDZHWBL21}, which includes 112 video clips of diverse scenes. The dataset features variable lighting conditions (sunny, cloudy, or nighttime), object sizes (diameters of 15 pixels or more), and densities (average object counts per frame above or below 150). Captured with a high-definition camera at 1920 $\times$ 1080 resolution and 25 frames per second (FPS), the dataset provides annotations for the trajectories of 20,800 individuals and 4.8 million heads. It divides these clips into 142 sequences of 300 frames each, allocating 82 sequences for training, 30 for validation, and 30 for testing. For evaluating crowd tracking algorithms, we use temporal mean average precision (T-mAP) to measure trajectory accuracy, incorporating thresholds (T-AP@0.10, T-AP@0.15, T-AP@0.20) and a 25-pixel accuracy threshold for validating track segments. These metrics, specified in the publication introducing \textsc{DroneCrowd}, are tailored for precise tracking assessment.

\subsection{Implementation Details}
% We implement our method based on the PyTorch framework. 
% To efficiently train FIDT~\cite{tmm/LiangXZZ23}, we choose the adaptive moment estimation (Adam)~\cite{corr/KingmaB14} optimizer. In training, the batch size is set to 16, and the crop size is set to 256. 
% % The VLP model BLIP2 is currently encapsulated in LAVIS\footnote{https://github.com/salesforce/LAVIS}. 
% To streamline operations, we directly feed cropped head images sized at 20 $\times$ 20 into the unified interface offered by LAVIS\footnote{https://github.com/salesforce/LAVIS} to acquire extracted image features. As we combine appearance and motion cues, we assign a weight of 0.9 to the parameter $\lambda$. This entire process is executed on the NVIDIA GeForce RTX 4090 platform.
We developed our method using the PyTorch framework. For efficient training of the FIDT model~\cite{tmm/LiangXZZ23}, we employ the Adam optimizer~\cite{corr/KingmaB14}. The training parameters include a batch size of 16 and a crop size of 256. To facilitate operations, cropped head images of 20 $\times$ 20 are fed directly into the LAVIS interface\footnote{https://github.com/salesforce/LAVIS} for feature extraction. We integrate appearance and motion cues by assigning a weight of 0.9 to the parameter $\lambda$. The entire training process is conducted on an NVIDIA GeForce RTX 4090 platform.

% \subsection{Comparison with state-of-the-art methods}
% 双栏表

\begin{table*}[t]
	\centering
	\caption{\small Tracking performances on \textsc{DroneCrowd}; average T-mAP, and T-AP at each threshold (T-AP$_{0.10}$, T-AP$_{0.15}$, and T-AP$_{0.20}$). MOT and DCT stands for Multi Object Tracking and Drone-based Crowd Tracking, respectively. The best results are highlighted in bold.}% 
	\footnotesize
	\setlength{\tabcolsep}{15pt}
	\begin{tabular}{lccccccc}
	\toprule
	{Method} & Venue & {MOT} & {DCT} & {T-mAP} & {T-AP$_{0.10}$} & {T-AP$_{0.15}$} & {T-AP$_{0.20}$}\\
	\midrule
 	MCNN~\cite{cvpr/ZhangZCGM16} & CVPR'16 & \Circle & \CIRCLE & 9.16 & 11.47 & 9.65 & 6.36\\
 	CSRNet~\cite{cvpr/LiZC18} & CVPR'18 & \Circle & \CIRCLE & 12.15 & 17.34 & 12.85 & 6.26\\
	CAN~\cite{cvpr/LiuSF19} & CVPR'19 & \Circle & \CIRCLE & 4.39 & 6.97 & 4.72 & 1.48\\
	DM-Count~\cite{nips/0001LSN20} & NeurIPS'20 & \Circle &\CIRCLE & 17.01 & 22.38 & 18.34 & 10.29\\
	STNNet~\cite{cvpr/WenDZHWBL21} & CVPR'21 & \CIRCLE & \CIRCLE & 32.50 & 35.45 & 33.99 & 28.05\\
 	StrongSORT~\cite{tmm/DuZSZSGM23} & TMM'23 & \CIRCLE & \Circle& 8.98 & 10.63 & 8.96 & 7.34\\
 	BoT-SORT~\cite{corr/abs-2206-14651} & arXiv'23 & \CIRCLE & \Circle& 13.60& 14.60& 13.63& 12.58\\
 	Deep-OC-SROT~\cite{icip/MaggiolinoACK23} & ICIP'23 & \CIRCLE & \Circle & 28.39 & 30.84 & 28.52 & 25.81\\
 	OC-SORT~\cite{cvpr/CaoPWKK23} & CVPR'23 & \CIRCLE & \Circle& 34.26 & 38.30 & 34.25 & 30.22\\
	\midrule
	DenseTrack (Ours) & & \CIRCLE & \CIRCLE & \textbf{39.44}& \textbf{47.48} & \textbf{39.88} & \textbf{30.95}\\
	\bottomrule
	\end{tabular}
	\label{tab:ExpSOTA}
\end{table*}%

\subsection{Comparison with State-of-the-Arts} 
Table~\ref{tab:ExpSOTA} presents a comparative analysis of tracking performance on \textsc{DroneCrowd}. STNNet~\cite{cvpr/WenDZHWBL21} relies solely on motion-based methods, its performance is suboptimal as it may misidentify individuals in close proximity. 
In contrast, DenseTrack integrates both motion and appearance, mitigating this issue. It performs exceptionally well, achieving the highest T-mAP score of 39.44, excelling across all thresholds.
Demonstrating outstanding tracking capability, especially in challenging environments, DenseTrack's strong performance at lower thresholds highlights effectiveness under less stringent conditions, while competitiveness at higher thresholds showcases reliability across diverse tracking scenarios.

\begin{table}[t]
 	\centering
 	\caption{\small Detection performances on \textsc{DroneCrowd}; The columns under ``Counting'' represent localization errors using only the counting network, while those under ``Tracking'' show errors refined through both the counting and tracking networks. The best results are highlighted in bold.}
 	\footnotesize
 	\setlength{\tabcolsep}{10pt}
 	\begin{tabular}{lcccc}
 	\toprule
 	\multirow{2}[2]{*}{Method} & \multicolumn{2}{c}{Couting} & \multicolumn{2}{c}{Tracking}\\
 	\cmidrule(lr){2-3} \cmidrule(lr){4-5} 
 	& MAE & RMSE & MAE & RMSE\\
 	\midrule
 	STNNet~\cite{cvpr/WenDZHWBL21} & \textbf{15.8} & \textbf{18.7} & 59.2 & 69.2\\
 	MPM~\cite{cvpr/HayashidaNB20} & 22.1 & 31.5&22.1 & 31.5\\
 	DenseTrack (Ours) & 20.3 & 21.4 & \textbf{19.2} & \textbf{29.0}\\
 	\bottomrule
 	\label{table:density}
 	\end{tabular}
\end{table}

\subsection{Ablation Study}
% In this section, we conduct some ablation studies to demonstrate the effectiveness of the components combined with our methods.
\subsubsection{Ablation Study on Density Localization}
Table~\ref{table:density} presents a comparison of direct counting-based human localization versus tracking-enhanced localization. When considering only counting, STNNet~\cite{cvpr/WenDZHWBL21} outperforms other methods with MAE of 15.8 and RMSE of 18.7. However, upon integration of tracking, STNNet experiences a significant increase in errors, with MAE of 59.2 and RMSE of 69.2. In contrast, while DenseTrack initially shows slightly higher errors in counting alone compared to STNNet, with MAE of 20.3 and RMSE of 21.4, it substantially improves localization accuracy with tracking adjustments, achieving MAE of 19.2 and RMSE of 29.0. 
This highlights DenseTrack's effectiveness in leveraging tracking information to improve localization accuracy, thereby outperforming STNNet in the tracking-enhanced scenario.

\begin{figure}[t]
 	\centering
 	\includegraphics[width = \linewidth]{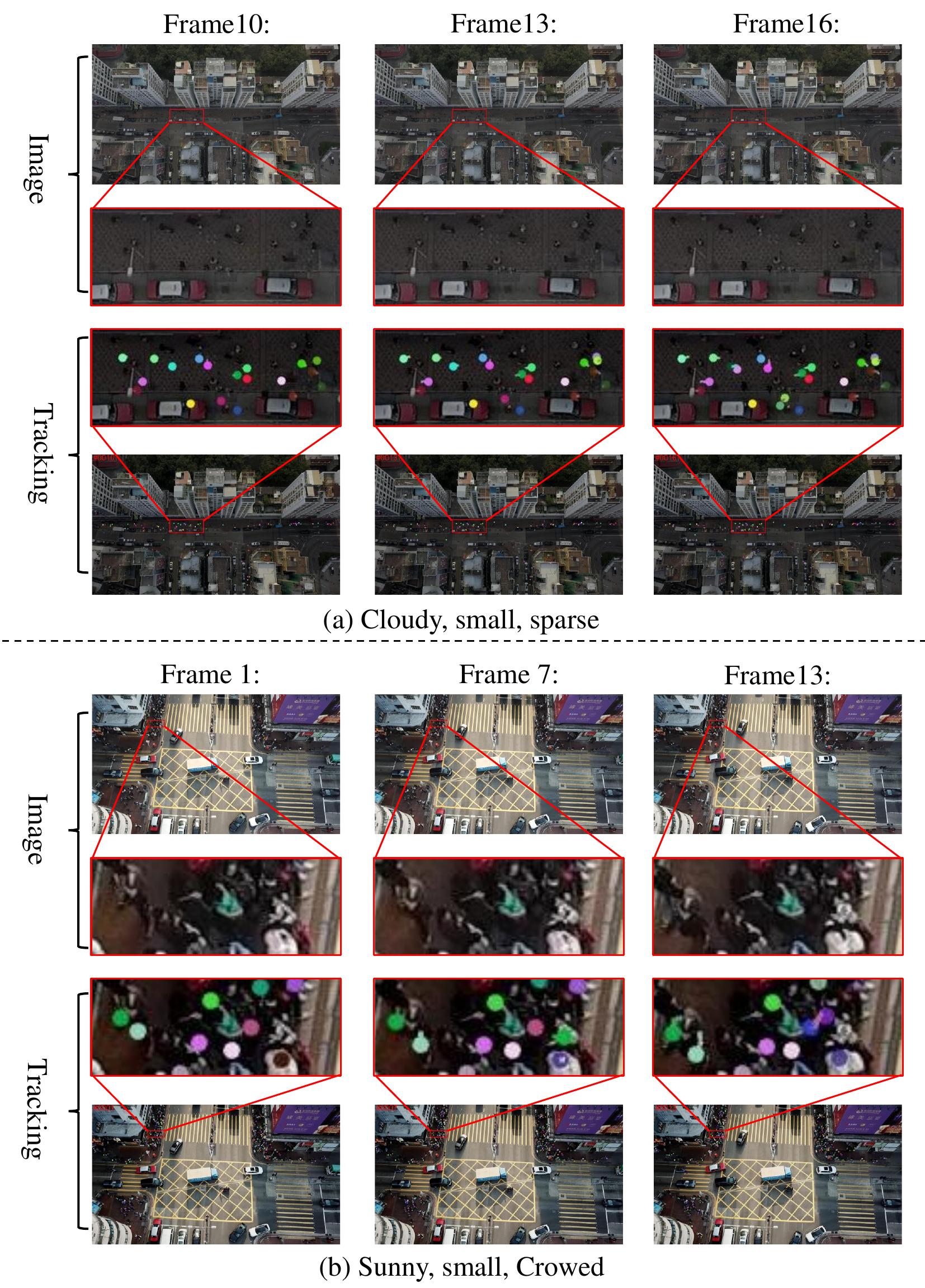}
 	\caption{\small Illustration of tracking under different conditions. (a) Sparse small objects in cloudy weather conditions. (b) Dense small objects in sunny weather conditions, with the same color representing the same individual.}
 	\label{fig3}
\end{figure}

\begin{table}[t]
	\centering
	\caption{\small Ablation studies investigate different factors influencing tracking performance, with each row depicting the impact of various solutions on tracking performance. ``Counting'' denotes tracking based solely on counting for localization and motion information tracking, ``Appearance'' represents appearance information, and ``HA'' stands for the Hungarian algorithm for matching. The best results are highlighted in bold.}% 
 	\footnotesize
	\begin{tabular}{cccccccc}
	\toprule
	Counting & Appearance & HA & T-mAP & T-AP$_{0.10}$ & T-AP$_{0.15}$ & T-AP$_{0.20}$\\ 
	\midrule
	 \CIRCLE & \Circle & \Circle& 2.90& 3.45 & 2.95& 2.29\\
	 \CIRCLE & \CIRCLE & \Circle & 37.46 & 45.59 & 37.80 & 28.99\\
	 \CIRCLE & \CIRCLE & \CIRCLE & \textbf{39.44}& \textbf{47.48} & \textbf{39.88} & \textbf{30.95}\\
	\bottomrule
	\end{tabular}
	\label{table3}
\end{table}

\begin{figure*}[t]
 	\centering
 	\includegraphics[width = 0.99\textwidth]{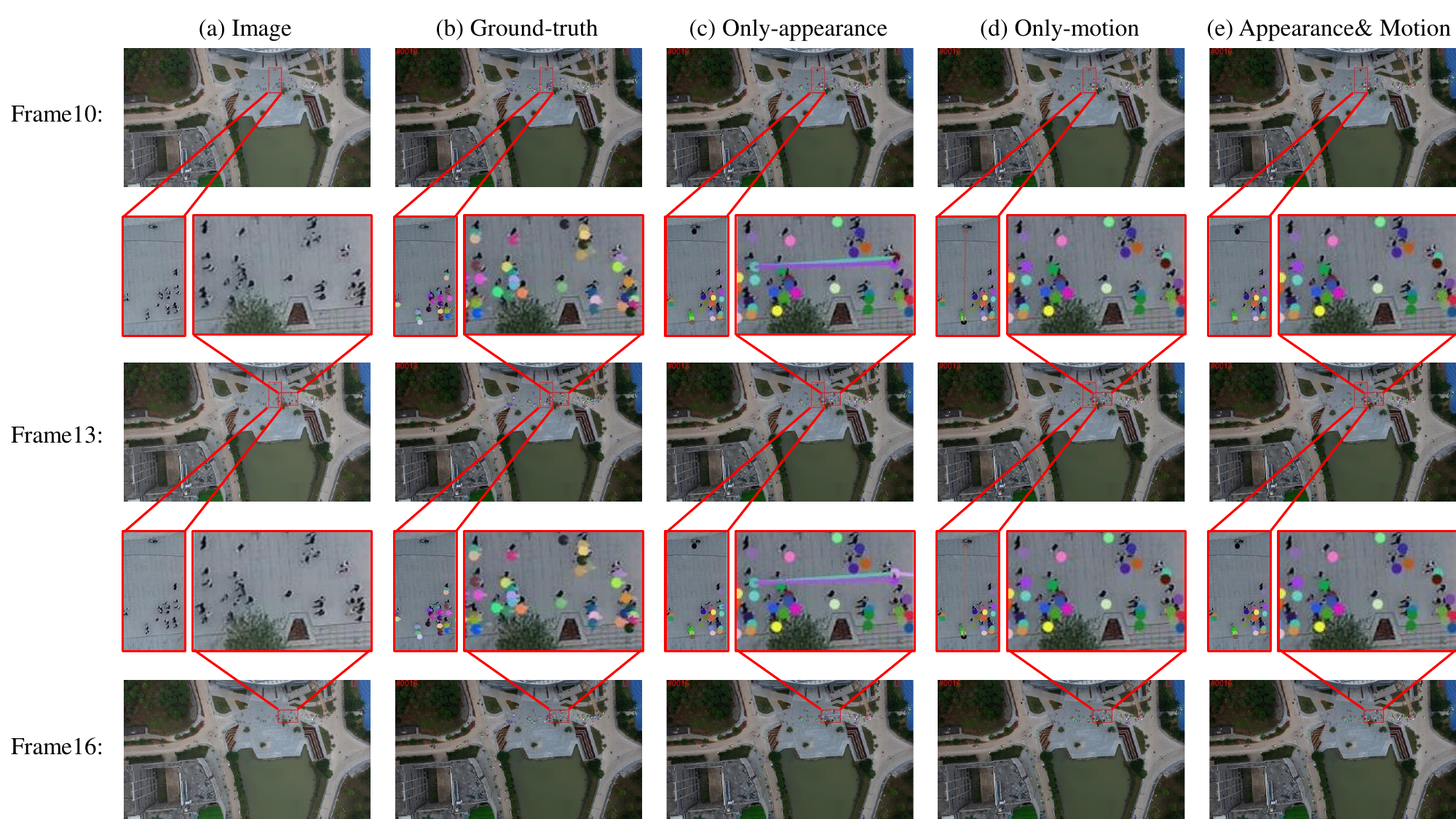}
 	\caption{\small Illustration of tracking performance using different strategies across frames 10, 13, and 16: (a) original aerial image, (b) ground-truth annotations, (c) tracking based solely on appearance, (d) tracking based solely on motion, and (e) tracking integrating appearance and motion. Insets magnify tracking results, showcasing the performance of each strategy.}
 	\label{fig4}
\end{figure*}

\subsubsection{Ablation Study on Various Factors Performance}
To evaluate the contribution of each component to the enhancement of tracking performance, we present the results of tracking effectiveness after omitting certain steps, as detailed in Table~\ref{table3}.

Specifically, the first row employs only counting-based localization and motion tracking, resulting in relatively low T-mAP (2.90) and T-AP at various thresholds (T-AP${0.10}$: 3.45, T-AP${0.15}$: 2.95, T-AP${0.20}$: 2.29). Introducing appearance information in the second row leads to significant improvements across all metrics, particularly with T-mAP increasing to 37.46, and notable enhancements in T-AP thresholds (T-AP${0.10}$: 45.59, T-AP${0.15}$: 37.80, T-AP${0.20}$: 28.99). However, the most significant performance boost is observed in the third row, where all factors are combined. Here, T-mAP reaches 39.44, and T-AP thresholds peak (T-AP${0.10}$: 47.48, T-AP${0.15}$: 39.88, T-AP$_{0.20}$: 30.95). These results underscore the critical role of considering appearance information and employing a matching algorithm for achieving optimal tracking performance in \textsc{DroneCrowd}.

\begin{table}[t]
	\centering
	\caption{\small Ablation studies evaluating tracking performance using different VLMs: CLIP, BLIP, and BLIP2. The best results are highlighted in bold.}
 	\footnotesize
 	\setlength{\tabcolsep}{10pt}
	\begin{tabular}{lccccc}
	\toprule
	Method & T-mAP & T-AP$_{0.10}$ & T-AP$_{0.15}$ & T-AP$_{0.20}$\\
	\midrule
	% ALBEF & & & &\\
	CLIP~\cite{Rad2021CLIP} & 39.33 & 47.25 & 39.64 & 31.12\\
	BLIP~\cite{Li2022BLIP} & 39.19 & 47.07 & 39.68 & 30.82\\
	BLIP2~\cite{Li2023BLIP2} & \textbf{39.44}& \textbf{47.48} & \textbf{39.88} & \textbf{30.95}\\
	\bottomrule
	\end{tabular}
 	\label{tab:4}
\end{table}

\subsubsection{Ablation Study on Visual Representation}
% In addition to BLIP, LAVIS also provides many VLP model interfaces for feature extraction, such as the newly proposed BLIP2, and CLIP.
Table~\ref{tab:4} presents the tracking performance of different visual-language models (VLMs), showcasing their effectiveness in improving tracking accuracy. While all methods exhibit notable performance, BLIP2~\cite{Li2023BLIP2} stands out as the top performer, achieving a T-mAP score of 39.44. This result underscores the efficacy of BLIP2 in enhancing tracking performance compared to other VLMs like CLIP~\cite{Rad2021CLIP} and BLIP~\cite{Li2022BLIP}. The consistent superiority of BLIP2 across various precision thresholds highlights its robustness and effectiveness in capturing intricate visual and language cues for more accurate tracking. This analysis suggests that BLIP2's architecture incorporates beneficial features that contribute to its superior performance, making it a promising choice for tracking tasks in diverse scenarios.

\begin{figure*}[t]
 	\centering
 	\includegraphics[width = 0.99\textwidth]{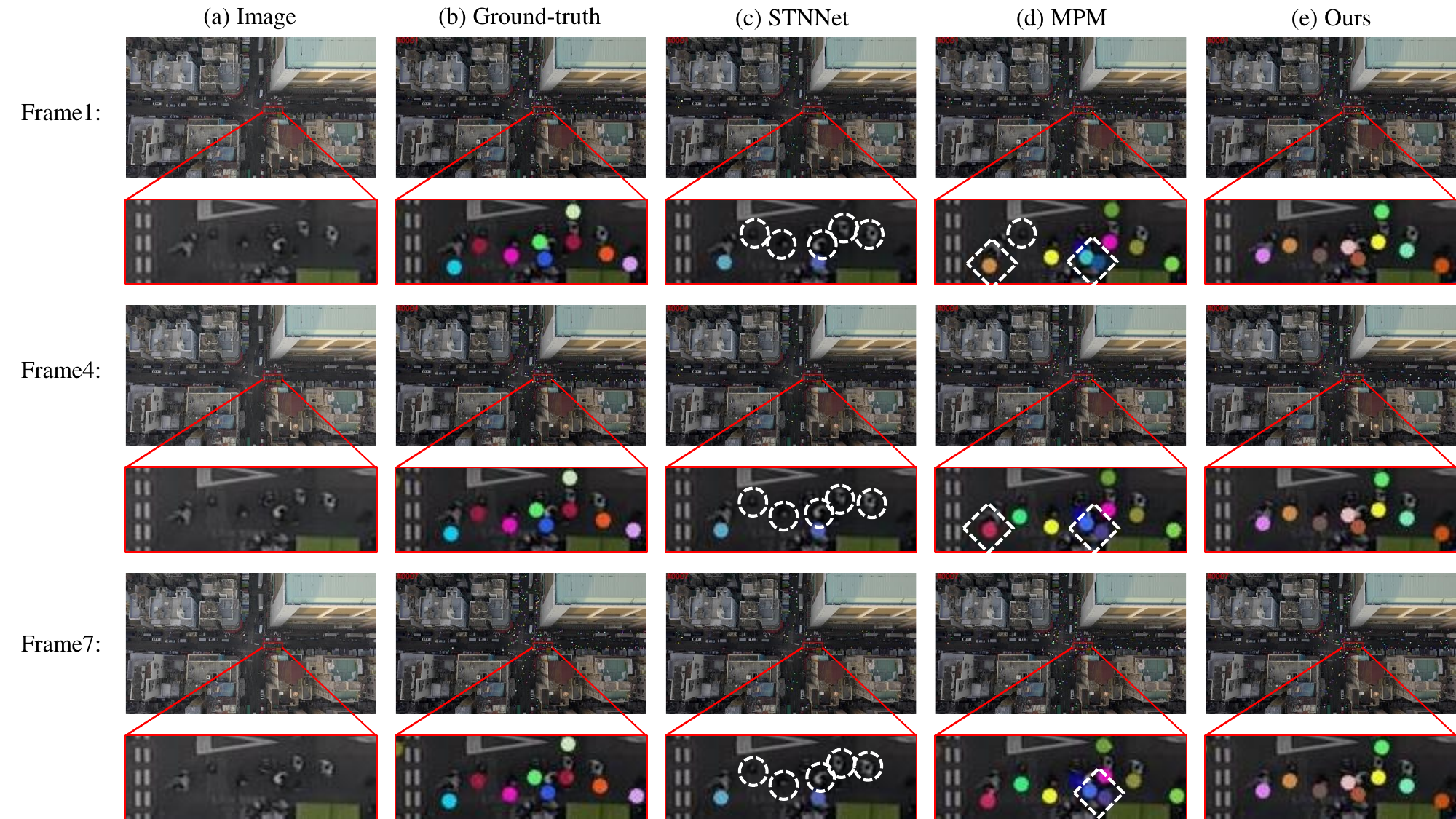}
 	\caption{\small Comparison of different tracking methods across frames 1, 4, and 7: (a) original surveillance footage, (b) ground-truth annotations, (c) tracking results from STNNet, (d) tracking results from MPM, and (e) our DenseTrack results. False negatives are marked with white dotted circles, and tracking switch errors with white rectangles. Insets provide a detailed view of tracking discrepancies, using consistent color coding to identify each individual.}
 	\label{fig5}
\end{figure*}

\subsection{Qualitative Analysis}
\subsubsection{Analysis of Tracking Performance in Varied Conditions}
Fig.~\ref{fig3} showcases the capability of DenseTrack to effectively manage complex tracking scenarios. 
DenseTrack reliably identifies and tracks individuals across various environmental challenges, maintaining its accuracy even under conditions of cloud cover and high crowd density. This performance demonstrates the robustness of integrating motion and appearance cues within DenseTrack, allowing for precise tracking that is largely unaffected by scene complexities. The framework's adeptness in such diverse conditions underscores its advanced design and suitability for varied aerial applications.

\subsubsection{Analysis of Different Tracking Strategies}
% Fig.~\ref{fig4} visually compares different tracking strategies to highlight the effectiveness of integrating appearance and motion information. The appearance-only strategy (\textit{cf.} Fig.~\ref{fig4}(c)) though it accurately identifies all individuals and tracks most correctly, suffers from errors over long distances. These are significantly reduced when motion information is included. The motion-only strategy (\textit{cf.} Fig.~\ref{fig4}(d)) avoids long-distance errors but tends to misidentify nearby targets. By combining both approaches, the integrated method (\textit{cf.} Fig.~\ref{fig4}(e)) effectively balances distance considerations, minimizes errors with proximal targets, and thereby achieves optimal tracking performance.
Figure~\ref{fig4} visually compares various tracking strategies, illustrating the benefits of integrating appearance and motion information. The appearance-only strategy (see Fig.\ref{fig4}(d)) minimizes long-distance errors but often misidentifies nearby targets. The integrated approach (see Fig.~\ref{fig4}(e)) successfully merges these strategies, effectively balancing distance considerations and minimizing proximity errors, resulting in optimal tracking performance.

\subsubsection{Analysis of Tracking Performance}
Fig.~\ref{fig5} offers an insightful comparative analysis, shedding light on the efficacy of our DenseTrack algorithm when juxtaposed with two prominent counterparts: STNNet~\cite{cvpr/WenDZHWBL21} and MPM~\cite{cvpr/HayashidaNB20}. Each snapshot within the figure unveils distinct facets of the localization challenges and tracking discrepancies inherent in these methodologies.
Examining STNNet's depiction (\textit{cf.} Fig.~\ref{fig5}(c)), significant localization errors are evident, highlighting the pivotal role of robust localization techniques in tracking precision. Conversely, the MPM-based approach (\textit{cf.} Fig.~\ref{fig5}(d)) shows some improvement but remains prone to occasional false detections.
In contrast, the DenseTrack method (\textit{cf.} Fig.~\ref{fig5}(e)) notably improves localization accuracy and tracking precision. Its ability to accurately identify and track individuals across various scenarios underscores its effectiveness in addressing complex tracking challenges.

\section{Conclusion and Discussion}
In this work, we present DenseTrack, a novel tracking-by-counting method that enhances drone-based crowd monitoring by integrating appearance and motion cues. We construct a cost matrix combining a density-aware appearance similarity matrix with a cross-frame motion distance matrix, and apply the Hungarian algorithm to achieve robust tracking outcomes. DenseTrack demonstrates competitive performance in crowded drone surveillance environments. To the best of our knowledge, this is the first implementation to synergistically use both appearance and motion information for drone-based crowd tracking.

\textbf{Limitations and Future Work.} Despite its strengths, DenseTrack is not fully optimized for all environmental conditions and tends to underperform in low-light or cloudy scenarios. Future research will focus on enhancing its adaptability and effectiveness in a wider array of challenging surveillance contexts.

\begin{acks}
This work was supported in part by the National Natural Science Foundation of China under Grant 62271361.
\end{acks}


% \clearpage
\bibliographystyle{ACM-Reference-Format}
\bibliography{Tracking}

% \newpage
\appendix
\begin{figure*}[t]
 	\centering
 	\includegraphics[width = 0.99\textwidth]{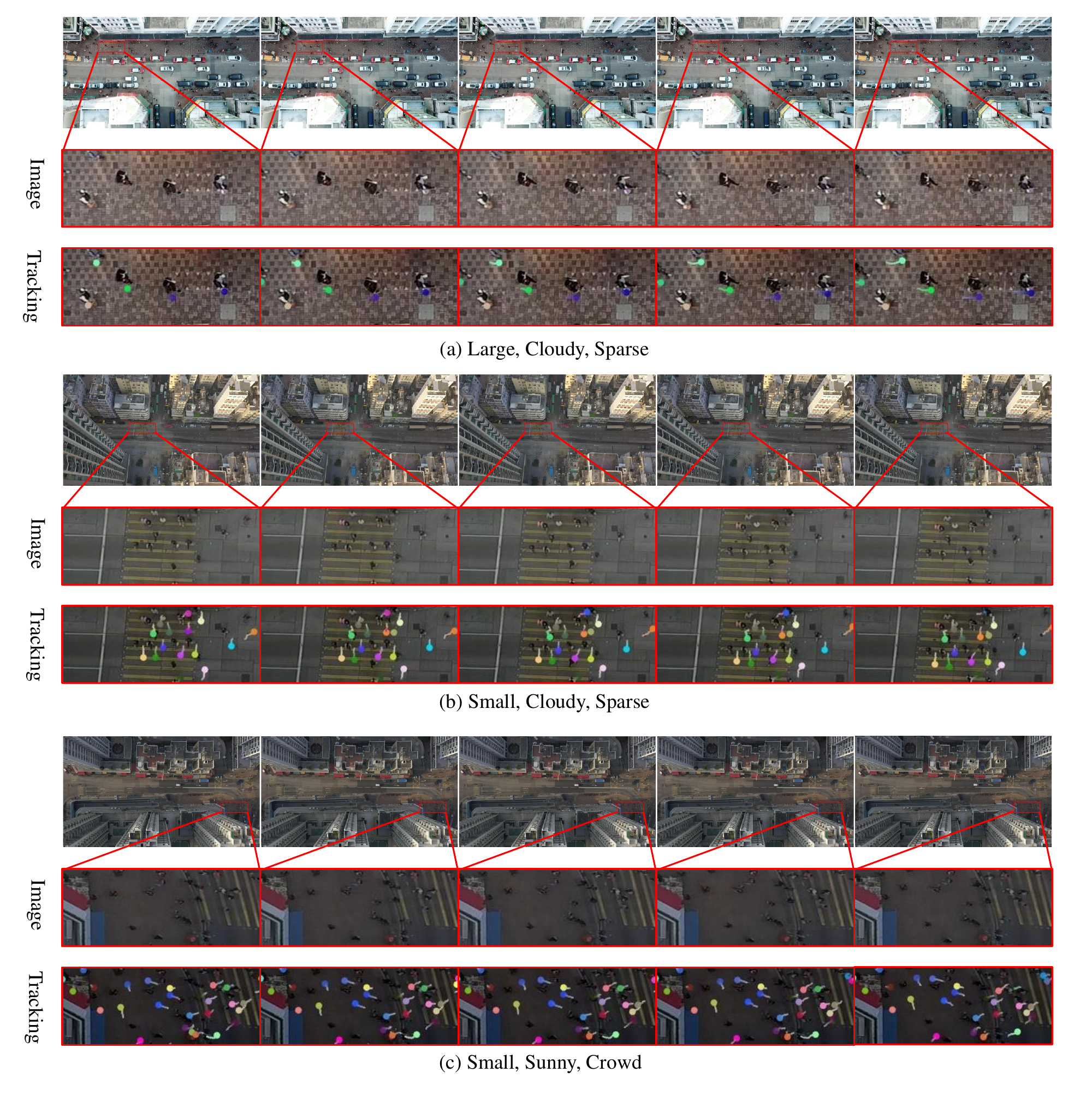}
 	\caption{\small Illustration of tracking under different scenarios. (a) Sparse large objects in cloudy weather scenarios. (b) Sparse small objects in cloudy weather scenarios. (c) Crowd small in sunny weather scenarios.}
 	\label{fig1a}
\end{figure*}
\section*{appendix}

\section{Ablation Study on Retrieval Method}
% When the same person is associated between frames, this paper uses diffusion to obtain the similarity matrix, which calculates the distance between the personal representations of two people. Therefore, this paper also uses other distance measures for calculations. 

Table~\ref{table:5} provides a comprehensive comparison of the impact of different retrieval methods on tracking performance. Across all evaluated distance metrics, Cosine, Euclidean, and Diffusion, we observe notable improvements in tracking accuracy. While each method demonstrates effectiveness, the diffusion retrieval method stands out with the highest T-mAP score of 39.44 and T-AP$_{0.10}$ score of 47.48. This signifies its superior performance in associating individuals across frames. 
The results indicate that leveraging appearance-based retrieval methods, especially through diffusion, notably improves tracking accuracy.
% \subsubsection{Ablation Study on Crow Flow Calculation}
% The optimal experimental outcomes are achieved using the Hungarian algorithm pro, which involves applying distance filtering followed by the Hungarian algorithm. It's conceivable that the performance enhancement is attributed to distance filtering rather than to the composite method itself. Therefore, experimenting solely with distance filtering reveals that the Hungarian algorithm pro method yields superior results. This indicates that employing distance as a criterion to constrain the matching scope, in conjunction with the Hungarian algorithm, significantly enhances tracking performance.

\begin{table}[h]
	\centering
	\caption{\small Ablation studies comparing the impact of distance measurement on the similarity matrix of appearances. Each row shows the performance using Cosine, Euclidean, and Diffusion distance. The best results are highlighted in bold.}
	\footnotesize
	\setlength{\tabcolsep}{9pt}
	\begin{tabular}{lcccc}
	\toprule
	Retrieval Method & T-mAP & T-AP$_{0.10}$ & T-AP$_{0.15}$ & T-AP$_{0.20}$\\
	\midrule
	Cosine & 39.32 & 47.02 & 39.88 & \textbf{31.05}\\
	Euclidean & 39.33 & 47.13 & \textbf{39.91} & 30.95\\
	% KL & & & &\\
	Diffusion~\cite{yang2019efficient} & \textbf{39.44} & \textbf{47.48} & {39.88} & {30.95}\\
	\bottomrule
	\end{tabular}
 	\label{table:5}
\end{table}

\section{Visualization of Tracking Results in Various Scenarios}
% 我们将不同场景钟的跟踪结果的可视化如图1所示。在多目标跟踪任务中，Sparse small objects是易于进行外观特征提取和运动偏移量预测的，因此也易于完成跟踪任务， 而Crowd small objects则恰好相反。因此， 这些结果表明，无论是易于跟踪的场景中还是不不易于跟踪的场景中，我们的方法都能展现出很好的跟踪性能。
% We present visualizations of tracking results in different scenarios as shown in Fig \ref{fig1}. In multi-object tracking tasks, Sparse large objects are easier to track due to their ease of extracting appearance features and predicting motion offsets. Conversely, Crowd small objects pose a greater tracking challenge because of the difficulty in extracting appearance features and the insufficient discriminative power of motion offsets, making tracking more difficult. Therefore, these results indicate that our method demonstrates excellent tracking performance in both easy-to-track and challenging scenarios.
We present visualizations of tracking results in different scenarios as shown in Fig.~\ref{fig1a}. In multi-object tracking tasks, Sparse large objects are easier to track due to their ease of extracting appearance features and predicting motion offsets. Conversely, Crowd small objects pose a greater tracking challenge due to the difficulty in extracting appearance features and the insufficient discriminative power of motion offsets, making tracking more difficult. Therefore, these results indicate that our method demonstrates excellent tracking performance in both easy-to-track and challenging scenarios.

\section{Comparison with the State-of-arts in Crowd Localization Performance}

Table~\ref{tab:6} provides a competitive analysis of the localization performance of various methods on \textsc{DroneCrowd}. While the success of our localization is influenced by~\cite{cvpr/WenDZHWBL21} and is not the focus of our research, localization remains a crucial task in object tracking, determining the accuracy of tracking results. Therefore, we still conducted relevant experiments to demonstrate the effectiveness of using density maps for object detection.Following the paper that introduced \textsc{DroneCrowd}, we evaluate the localization performance of crowds using the L\_AP score. L\_mAP represents the average of L\_AP over different distance thresholds (1, 2, $\cdots$, 25 pixels). A smaller L\_AP distance threshold implies a stricter requirement for precision in localization, while a higher L\_AP value indicates better performance.

\begin{table}[h]
	\centering
	\caption{\small Localization performances on \textsc{DroneCrowd}; average L-mAP, and L-AP at each threshold (L-AP$_{10}$, L-AP$_{15}$, and L-AP$_{20}$). MOT and DCT stands for Multi Object Tracking and Drone-based Crowd Tracking, respectively. The best results are highlighted in bold.}% 
	\footnotesize
	\begin{tabular}{lcccccc}
	\toprule
	{Method} & {MOT} & {DCT} & {L-mAP} & {L-AP$_{10}$} & {L-AP$_{15}$} & {L-AP$_{20}$}\\
	\midrule
 	MCNN~\cite{cvpr/ZhangZCGM16} & \Circle & \CIRCLE & 9.05 & 9.81 & 11.81 & 12.83\\
	CAN~\cite{cvpr/LiuSF19} & \Circle & \CIRCLE & 11.12 & 8.94 & 15.22 & 18.27\\
 CSRNet~\cite{cvpr/LiZC18} & \Circle & \CIRCLE & 14.40 & 15.13 & 19.17 & 21.16\\
	DM-Count~\cite{nips/0001LSN20} & \Circle & \CIRCLE & 18.17 & 17.90 & 25.32 & 27.59\\
	STNNet~\cite{cvpr/WenDZHWBL21} & \CIRCLE & \CIRCLE & 40.45 & 42.75 & 50.98 & \textbf{55.77}\\
	DenseTrack (Ours) & \CIRCLE & \CIRCLE & \textbf{43.52} & \textbf{47.75} & \textbf{52.21} & 54.71\\
	\bottomrule
	\label{tab:6}
	\end{tabular}
\end{table}

It is noteworthy that our method exhibits a more significant improvement in $L\_AP$ under stricter thresholds (compared to STNNet~\cite{cvpr/WenDZHWBL21}, $L\_AP_{10}$ increases from 42.75\% to 47.75\%). This suggests that our method is more precise in detecting objects, which is highly beneficial for enhancing tracking performance. Furthermore, our method demonstrates improvements in overall metrics as well.

\section{Comparison of Motion Weight}
% 在帧间关联阶段，代价矩阵是由相似度矩阵与归一化后的距离矩阵加权和得到，如式1所示。
In the inter-frame association stage, the cost matrix $A^C_{i, i+1}$ is obtained by the weighted sum of the similarity matrix $A^S_{i, i+1}$ and the normalized distance matrix $\hat{A}^D_{i, i+1}$, as shown in Eq.~\eqref{eq16}.
\begin{equation}
 A^C_{i, i+1} = \left(-\lambda \right) \hat{A}^D_{i, i+1} + \left(1- \lambda \right) A^S_{i, i+1}.
 \label{eq16}
\end{equation}

\begin{figure}[h]
 	\centering
 	\includegraphics[width = 0.9\linewidth]{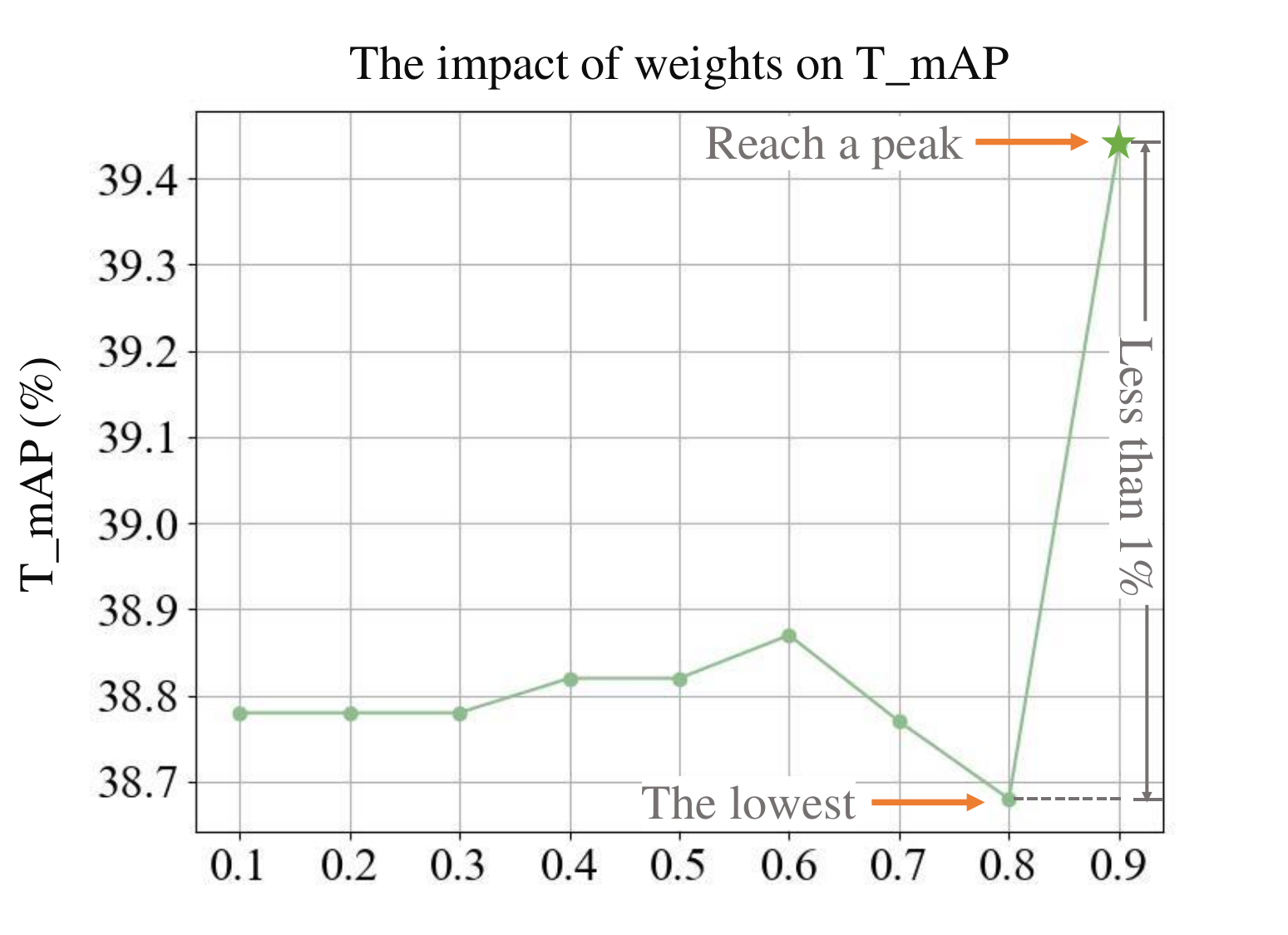}
 	\caption{\small Tracking performances under different weights on \textsc{DroneCrowd}}
 	\label{fig:weight}
\end{figure}

%图 ~\ref{fig:weight}展示了$\lamda$的不同取值对于跟踪性能的影响。在DenseTrack中，$\lamda$的取值对跟踪性能的影响并不大，以T_mAP为例，最高为39.44%， 最低为38.68%，差距不足1%。
Fig.~\ref{fig:weight} illustrates the impact of different values of $\lambda$ on tracking performance. In DenseTrack, the choice of $\lambda$ does not significantly affect tracking performance. For instance, regarding T\_mAP, it ranges from a maximum of 39.44\% to a minimum of 38.68\%, with a difference of less than 1\%.

\begin{figure*}
	\centering
	\includegraphics[width = 0.99\textwidth]{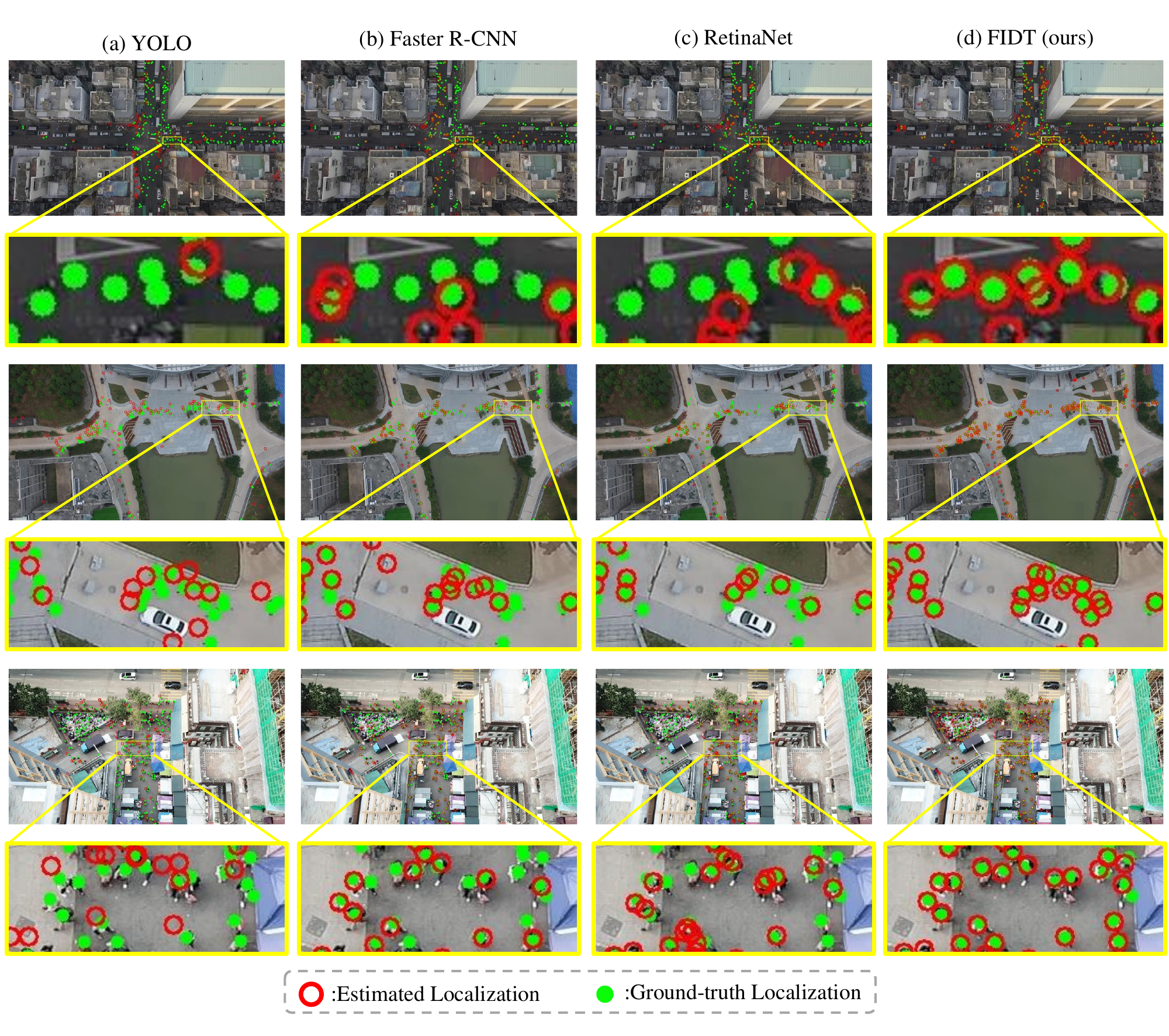}
	\caption{\small Illustration of localization under different detector. (a) Use YOLOv8 to detect objects. (b) Use Faster R-CNN to detect objects. (c) Use RetinaNet to detect objects. (d) Use density map (FIDT) to detect objects. }
	\label{fig:8}
\end{figure*}

\section{Comparsion with Other Detector in Crowd Localization Performance}
% Tab.~\ref{tab:7}展示了不同检测器在\textsc{DroneCrowd}上的定位性能。由于\textsc{DroneCrowd}是由无人机在高空俯视视角下拍摄的，因此目标与背景之间的差别并不明显，这也影响了常用的检测器在定位性能上的表现。
Tab.~\ref{tab:7} displays the localization performance of different detectors on \textsc{DroneCrowd}. Since \textsc{DroneCrowd} is captured from a high-altitude overhead perspective by drones, the distinction between objects and background is not very pronounced, which also affects the performance of commonly used detectors in terms of localization.

\begin{table}[h]
	\centering
	\caption{\small Localization performances of different detector on \textsc{DroneCrowd}; average L-mAP, and L-AP at each threshold (L-AP$_{10}$, L-AP$_{15}$, and L-AP$_{20}$);}% 
	\footnotesize
	\setlength{\tabcolsep}{10pt}
	\begin{tabular}{lcccccc}
	\toprule
	{Method} & {L-mAP} & {L-AP$_{10}$} & {T-AP$_{15}$} & {T-AP$_{20}$}\\
	\midrule
	YOLOv8~\cite{corr/abs-2310-01641} & 6.62 & 1.37 & 7.16 & 14.26\\
	Faster R-CNN~\cite{nips/RenHGS15} & 22.39 & 24.35 & 26.68 & 28.27\\
	RetinaNet~\cite{pami/LinGGHD20} & 22.63 & 24.28 & 28.37 & 30.78\\
	FIDT (ours) & \textbf{43.55}& \textbf{47.77} & \textbf{52.24} & \textbf{54.77}\\
	\bottomrule
	\end{tabular}
	\label{tab:7}
\end{table}

%DenseTrack受liang等人工作的启发，在目标定位阶段将原本用于人群计数的密度图用于人群定位。结果表明，这种做法可以有效地提高目标检测的准确性，便于后续的跟踪工作。
Inspired by the work~\cite{tmm/LiangXZZ23}, DenseTrack utilizes density maps originally designed for crowd counting in the object localization stage. The results indicate that this approach effectively enhances the accuracy of object detection, facilitating subsequent tracking tasks.
%% table3

\section{Visualization of Various Detector}
% 尽管定位工作并不是DenseTrack的重点任务，但是目标定位的准确与否直接影响到了外观特征提取结果的好坏，进而影响跟踪的结果。同时，Detection-based tracking范式的跟踪结果也是高度依赖目标检测的准确性的。因此，人群定位是跟踪的重要步骤。
While localization is not the primary focus of DenseTrack, the accuracy of object localization directly impacts the quality of appearance feature extraction, thereby influencing tracking results. Additionally, the tracking outcomes of the Detection-based tracking paradigm heavily rely on the accuracy of object detection. Therefore, crowd localization is a crucial step in tracking.

% Fig ~\ref{fig:8}进一步的展示了YOLO，Faster R-CNN， RetinaNet以及我们使用的FIDT在定位性能上的区别。可视化结果表明，其他的检测器在检测密集人群时会存在漏检的情况，而使用密度图进行检测则可以减少这类情况的发生。同时，使用密度图进行检测可以使定位更加准确，因此也更加利于后续的外观提取工作。
Fig.~\ref{fig:8} further demonstrates the differences in localization performance among YOLOv8~\cite{corr/abs-2310-01641}, Faster R-CNN~\cite{nips/RenHGS15}, RetinaNet~\cite{pami/LinGGHD20}, and the FIDT~\cite{tmm/LiangXZZ23} we employ. The visualization results indicate that other detectors may experience missed detections when detecting dense crowds, whereas using density maps for detection can reduce the occurrence of such cases. Additionally, employing density maps for detection can enhance localization accuracy, thereby facilitating subsequent appearance extraction tasks.

\end{document}